\pdfoutput=1
\documentclass[11pt]{article}

\usepackage[]{acl}

\usepackage{times}
\usepackage{latexsym}
\usepackage{multirow}

\usepackage[T1]{fontenc}
\usepackage[utf8]{inputenc}

\usepackage{microtype}

\usepackage{inconsolata}

\newcommand{\model}{\textbf{\textsc{ToXCL}}}
\usepackage{graphicx}
\usepackage{booktabs}
\usepackage{colortbl}
\usepackage[ruled,linesnumbered]{algorithm2e}
\usepackage{algorithmic}

\title{ToXCL: A Unified Framework for Toxic Speech \\ Detection and Explanation}

\author{
    Nhat M. Hoang\textsuperscript{\rm 1}\thanks{~~Equal contribution}, Xuan Long Do\textsuperscript{\rm 2,3}\footnotemark[1], Duc Anh Do\textsuperscript{\rm 1}, Duc Anh Vu\textsuperscript{\rm 1}, Luu Anh Tuan\textsuperscript{\rm 1} \thanks{~~Corresponding author.}\\
    \textsuperscript{\rm 1}Nanyang Technological University, Singapore,\\
    \textsuperscript{\rm 2}National University of Singapore, \\ 
    \textsuperscript{\rm 3}Institute for Infocomm Research (I$^2$R), A*STAR \\
    \texttt{\{nhat005, ducanh003, ducanh001\}@e.ntu.edu.sg} \\
    \texttt{xuanlong.do@u.nus.edu}, \texttt{anhtuan.luu@ntu.edu.sg} \\
}

\begin{document}
\maketitle
\begin{abstract}
The proliferation of online toxic speech is a pertinent problem posing threats to demographic groups. While explicit toxic speech contains offensive lexical signals, implicit one consists of coded or indirect language. Therefore, it is crucial for models not only to detect implicit toxic speech but also to explain its toxicity. This draws a unique need for unified frameworks that can effectively detect and explain implicit toxic speech. Prior works mainly formulated the task of toxic speech detection and explanation as a text generation problem. Nonetheless, models trained using this strategy can be prone to suffer from the consequent error propagation problem. Moreover, our experiments reveal that the detection results of such models are much lower than those that focus only on the detection task. To bridge these gaps, we introduce \model{}\footnote{\url{https://github.com/NhatHoang2002/ToXCL}}, a unified framework for the detection and explanation of implicit toxic speech. Our model consists of three modules: a \emph{(i) Target Group Generator} to generate the targeted demographic group(s) of a given post; an \emph{(ii) Encoder-Decoder Model} in which the encoder focuses on detecting implicit toxic speech and is boosted by a \emph{(iii) Teacher Classifier} via knowledge distillation, and the decoder generates the necessary explanation. \model{} achieves new state-of-the-art effectiveness, and outperforms baselines significantly. 
% Code is available at: \url{https://github.com/NhatHoang2002/ToXCL}.

\end{abstract}

\section{Introduction}
\label{sec:intro}

\emph{\textcolor{red}{\textbf{Warning:} This paper discusses and contains content that can be offensive or upsetting.}}

\noindent While social media has dramatically expanded democratic participation in public discourse, they have also become a widely recognized platform for the dissemination of toxic speech \cite{hatexplain, elsherief-etal-2021-latent, yu2022hate}. Online toxic speech, therefore, is prevalent and can lead the victims to serious consequences \cite{olteanu2018effect, farrell2019exploring}. For this reason, the development of toxic detection tools has received growing attention in recent years \cite{hutto2014vader, ribeiro2018characterizing, balkir2022necessity,JAHAN2023126232}. 

\begin{figure}[t!]
    \includegraphics[width=\linewidth]{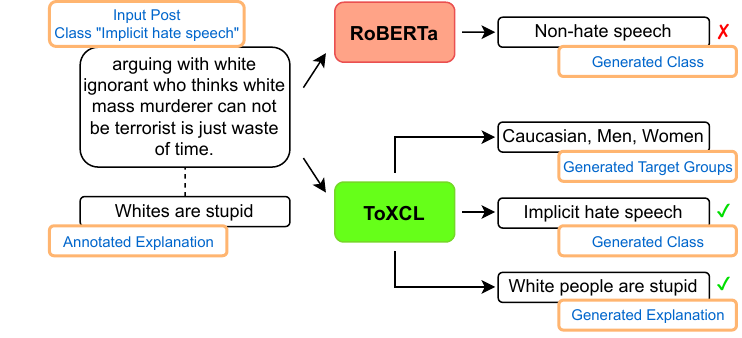}
    \caption{\small{A sample input post and its ground truth explanation from the test set of \texttt{Implicit Hate Corpus} \cite{elsherief-etal-2021-latent}. The input post is fed into both models. The baseline model, RoBERTa, fails to detect the implicit toxic speech while our proposed \model{}  model successfully detects it and generates a toxic explanation that closely matches the ground truth explanation.}}
    \label{fiq:introduction-example}
\end{figure}

Toxic speech can generally be categorized as either \textit{explicit} or \textit{implicit}. Explicit toxic speech contains direct offensive language targeting individuals or groups \cite{nockleyby2000hate} and has been extensively studied \cite{schmidt-wiegand-2017-survey, jahan2021systematic}. On the other hand, implicit toxic speech presents a more challenging detection task as it relies on stereotypes and indirect language \cite{elsherief-etal-2021-latent} (see Fig.~\ref{fiq:introduction-example}) and has received limited attention. Moreover, given the absence of explicit offensive words or cues, it is crucial for AI models not only to detect implicit toxic speech but also to provide explanations for its toxic nature \cite{sridhar-yang-2022-explaining}. The act of explanation serves practical purposes in various real-life applications, including improving human-machine interactions and building trustworthy AI systems \cite{ribeiro2016why, Dosilovic2018ExplainableAI, bai2022training}.

These applications, therefore, pose a need for unified systems that can effectively detect implicit toxic speech and explain its toxicity. However, previous works have mainly focused on a hybrid approach that combines detection and explanation tasks into a single text generation problem. For example, \citet{sap-etal-2020-social} proposed concatenating the toxic speech label and explanation as the target output, \citet{token} and \citet{huang2022chain} extended this approach by incorporating additional data such as target group(s) or social norms. Unfortunately, these hybrid approaches can introduce error propagation problems \cite{wu-etal-2018-beyond}, possibly due to differences in training objectives (see Sec.~\ref{sssec:teacher-classifier}). Consequently, models formulated in this manner tend to have much lower detection scores compared to models that focus solely on detection, as evidenced by our experimentation results (Sec.~\ref{sec:experimentation}). Another simple approach is building a modular-based system separating the detection module and the explanation generation module. However, in reality, this kind of framework is computationally expensive to train, store and deploy as it has multiple components.

\begin{figure*}[t!]
\centering
    \includegraphics[width=\textwidth]{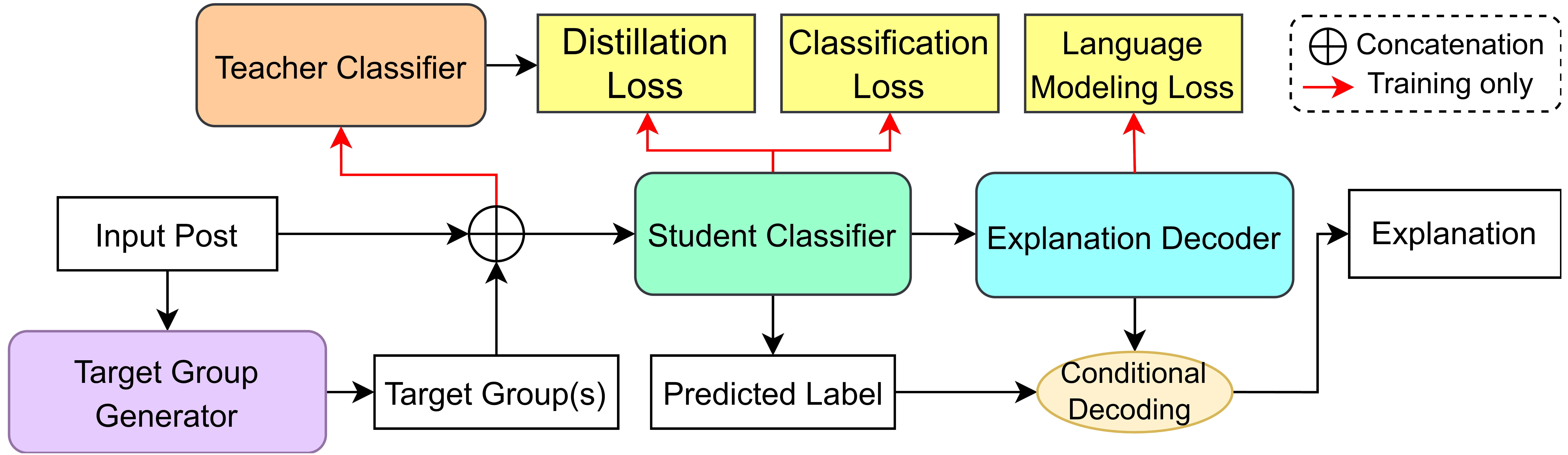}
    \caption{\small{An overview of our proposed \model{}. It consists of three modules: a \emph{(i) Target Group Generator} generates the target group(s) of the input post; an \emph{(ii) Encoder-Decoder Model} whose encoder focuses on implicit toxic speech detection whilst its decoder aims to generate necessary toxic explanation; a \emph{(iii) Teacher Classifier} to distil the knowledge to the classifier encoder.}}
\label{fiq:overview-framework}
\end{figure*}

To bridge these gaps in detecting and explaining implicit toxic speech, we propose a unified framework \model{} consisting of three modules (Fig.~\ref{fiq:overview-framework}). Our approach is motivated by the findings that modeling the minority target groups associated with toxic speech can potentially improve the performance of both implicit toxic detection and explanation tasks \cite{elsherief2018hate, token, huang2022chain}. To achieve this, we build a \emph{ Target Group Generator} as our first module to generate the target minority group(s) based on the input post. The generated target group(s) and the post are then input into an \emph{ Encoder-Decoder Model} whose encoder detects the speech, and the decoder outputs the necessary toxic explanation. To enhance the detection performance of our encoder, we incorporate a strong \emph{Teacher Classifier} that utilizes the \emph{teacher forcing} technique during training to distill knowledge to our encoder classifier. Finally, we introduce a \emph{Conditional Decoding Constraint} to enhance the explanation ability of the decoder during inference. Our model achieves state-of-the-art performance on the \texttt{Implicit Hate Corpus} (IHC) \cite{elsherief-etal-2021-latent} and \texttt{Social Bias Inference Corpus} (SBIC) \cite{sap-etal-2020-social} in the task of implicit hate speech detection and explanation, outperforming baselines. Our contributions are as follows:

\emph{(i)} We present a unified framework for the detection and explanation of implicit toxic speech. To the best of our knowledge, our work represents a pioneering effort in integrating both tasks into an encoder-decoder model to avoid the error propagation problem while maintaining the competitive performance on both tasks parameter-efficiently.

\emph{(ii)} We propose to generate target groups for the toxic speech detection and explanation model through the utilization of an encoder-decoder model, thereby distinguishing our approach from previous methods (see Sec.~\ref{sssec:tg-generator}). We also introduce several techniques to enhance the performance of our model: (1) joint training among the tasks to make the detection and explanation model end-to-end; (2) using teacher forcing to train the encoder; (3) a simple Conditional Decoding Constraint during the inference to avoid generating unnecessary explanation.

\emph{(iii)} We set up new strong state-of-the-art results in the task of implicit toxic speech detection and explanation tasks in two widely used benchmarks \texttt{Implicit Hate Corpus} (IHC) and \texttt{Social Bias Inference Corpus} (SBIC).

\emph{(iv)} We conduct a thorough analysis to demonstrate the effectiveness of our architectural design. We will open-source our model to inspire and facilitate future research.

%%%%%%%%%%%%%%%%%%%%%%%%%%%%%%%%%%%%%%%%%%%%%%%%%%%%%%%%%%%%%%%%%%%%%%%%%%%%%%%%%%%%%%
%%%%%%%%%%%%%%%%%%%%%%%%%%%%%%%%%%%%%%%%%%%%%%%%%%%%%%%%%%%%%%%%%%%%%%%%%%%%%%%%%%%%%%

\section{Related Work}
\label{sec:related-work}

\subsection{Toxic Speech Detection \& Explanation}
\label{ssec:related-toxic-work}
Early studies on identifying toxic speech relied on linguistic rule-based approaches \cite{6406271, hutto2014vader, gitari2015lexicon, wiegand-etal-2018-inducing}. However, these methods, which use predetermined lexical and syntactic rules, struggle to detect implicit toxic speech without explicit vulgarities \cite{breitfeller-etal-2019-finding, 10.1371/journal.pone.0221152}. Recent frameworks based on transformer architecture \cite{attentionisallyouneed} have made progress in detecting toxic speech \cite{basile-etal-2019-semeval, tran-etal-2020-habertor, kennedy-etal-2020-contextualizing}. However, detecting implicit toxic speech remains challenging despite attempts to improve performance on this task \cite{vidgen-etal-2019-challenges, caselli-etal-2020-feel, caselli-etal-2021-hatebert, kim-etal-2022-Generalizable}. The issue of explaining why a text is toxic has received even more limited attention, with some studies focusing solely on explaining implicit toxic speech \cite{elsherief-etal-2021-latent, sridhar-yang-2022-explaining}. Another few studies have addressed both implicit toxic speech detection and explanation \cite{sap-etal-2020-social, token, huang2022chain}, often formulating them as text generation tasks, possibly leading to error propagation and lower detection scores compared to detection-only models.

\subsection{Knowledge Distillation}
\label{ssec:related-knowledge-distil}
Knowledge distillation \cite{hinton2015distilling} is a technique that enables a smaller student model to learn from a larger teacher model by transferring knowledge. It has proven effective in improving performance, reducing computational requirements, and increasing efficiency in the field of Computer Vision \cite{Gou_2021}. Recently, researchers have explored applying knowledge distillation in Natural Language Processing. For example, \citet{https://doi.org/10.48550/arxiv.2012.07335} used a contrastive approach to align the intermediate layer outputs of the teacher and student models. \citet{https://doi.org/10.48550/arxiv.1908.08962} extensively studied the interaction between pre-training, distillation, and fine-tuning, demonstrating the effectiveness of pre-trained distillation in tasks like sentiment analysis. Additionally, \citet{clark-etal-2019-bam} trained a multitasking network by ensembling multiple single-task teachers. In our work, we distill the knowledge from a teacher classifier to our model's classifier (the student classifier), optimizing the Kullback-Leibler distance \cite{csiszar1975divergence} between soft labels.

%%%%%%%%%%%%%%%%%%%%%%%%%%%%%%%%%%%%%%%%%%%%%%%%%%%%%%%%%%%%%%%%%%%%%%%%%%%%%%%%%%%%%%
%%%%%%%%%%%%%%%%%%%%%%%%%%%%%%%%%%%%%%%%%%%%%%%%%%%%%%%%%%%%%%%%%%%%%%%%%%%%%%%%%%%%%%

\section{Methodology}
\label{sec:method}

\subsection{Problem Formulation}
\label{ssec:problem-definition}
The task of implicit toxic speech detection can be formulated as a binary classification problem while the explanation generation task can be regarded as a text generation problem. Each data instance ${\langle IP, Y, E \rangle}$ consists of an input post $IP$, a binary class label $Y$ ($1$ for toxic speech, $0$ for non-toxic speech), and a corresponding explanation $E$ (\texttt{[None]} for non-toxic speech). The models then take $IP$ as the input and learn to output $Y$ and $E$.

\subsection{\model{} Framework}
\label{ssec:toxcl}
Figure~\ref{fiq:overview-framework} shows an overview of our proposed \model{}, consisting of three modules: \emph{(i) Target Group Generator}; \emph{(ii) Encoder-Decoder Model}; \emph{(iii) Teacher Classifier}. The details of each module are presented below.

\subsubsection{Target Group Generator (TG)}
\label{sssec:tg-generator}
Current toxic speech detection systems often overlook the nuances of toxic speech, which can be better addressed by modeling the minority target groups associated with it \cite{elsherief2018hate}. Incorporating target group information has the potential to improve the accuracy of toxic speech detection and enable the generation of high-quality hate speech explanations \cite{huang2022chain}. Therefore, we propose using a transformer-based encoder-decoder model \cite{2020t5} to generate target minority groups for given posts, treating the task as a text generation problem rather than a multi-label classification task. This approach provides two advantages over classification models. Firstly, it leverages powerful pre-trained encoder-decoder models, enhancing the model's capabilities. Secondly, text generation models are more generalizable, as they are not restricted to a fixed number of target groups, allowing for greater flexibility in handling diverse scenarios.

After generating target groups $G$ based on an input post $IP$, $G$ and $IP$ are concatenated as \texttt{"Target:\{G\} Post:\{IP\}"} and serve as the input for our \model{}. The experimental details of the TG module are presented in Section~\ref{ssec:exp-tg-generator}.

\subsubsection{Encoder-Decoder Model}
\label{sssec:encoder-decoder-model}
Toxic speech detection and toxicity explanation are two tasks that have received increasing attention, and while researchers have made significant progress in separately solving each problem, addressing them together has received limited attention \cite{sap-etal-2020-social, token, huang2022chain}. However, these two tasks are strongly correlated, and the explanation of the post can potentially help the systems to detect toxic speech \cite{token}. Conversely, the toxicity explanation is sometimes only necessary when the post is detected as toxic. Typically, \citet{token, huang2022chain} formulate both tasks as a single text generation task, which has some critical shortcomings as discussed in Section~\ref{sec:intro}. Therefore, in this work, we propose a novel architectural design on top of a pre-trained encoder-decoder model. The encoder addresses the toxic speech detection task, while the decoder generates a toxicity explanation if the post is detected as toxic. The details of both the encoder and decoder components are introduced below.

\paragraph{$\bullet$ Encoder Classifier (CL)}
To enable the implicit toxicity detection capability, we build a binary classifier head on top of the encoder of a pre-trained encoder-decoder model. This head includes a linear layer that takes the average of token embeddings from the encoder's last hidden state as input, followed by a softmax layer \cite{goodfellow2016deep}. To optimize the performance, both the encoder and the newly added classifier head are trained together using a binary cross-entropy loss:

\begin{equation}
\label{eq:cls-loss} 
\mathcal{L}_{cls} = -\frac{1}{N} \sum_{i}^{}\sum_{j \in \{0,1\}}^{} y_{ij}\log (p_{ij})
\end{equation}

\noindent in which $y_{i0}, y_{i1} \in \{0,1\}$, $p_{i0}, p_{i1} \in [0,1]$ and $p_{i0} + p_{i1} = 1$.

\paragraph{$\bullet$ Explanation Decoder (ToX)}
Recognizing the importance of generating explanations for implicit toxic speech and its potential impact on various applications, we utilize the decoder of our pre-trained encoder-decoder model to generate the explanation. To optimize its performance, the decoder is fine-tuned using a Causal Language Modeling (CLM) loss:

\begin{equation}
\label{eq:clm-loss} 
\mathcal{L}_{clm} = \sum_{i}^{}\log(P(e_i|e_{i-k},...,e_{i-1};\theta))
\end{equation}

\noindent in which $E = \{e_1,e_2,...,e_n\}$ is the set of tokens of the explanation, and $k$ is the size of the window.

Finally, we train the encoder-decoder model for the task of toxic speech detection and explanation by joining the two losses:

\begin{equation}\label{eq:encoder-decoder-loss} 
\mathcal{L}_{xcl} = \alpha\mathcal{L}_{cls} + \beta\mathcal{L}_{clm} 
\end{equation}

\noindent in which $\alpha, \beta$ are the contribution weights. 

\subsubsection{Teacher Classifier (TC)}
\label{sssec:teacher-classifier}
Since the open-sourced encoder-decoder models are commonly pre-trained on a diverse range of tasks and these tasks might not solely focus on learning strong representations from their encoders, these encoders may not exhibit the same strength as pre-trained encoder-based models such as BERT \cite{devlin-etal-2019-bert} or RoBERTa \cite{liu2019roberta} for classification tasks. Motivated by \citet{hinton2015distilling}, we propose to use knowledge distillation to transfer knowledge from a strong encoder-based model (\emph{Teacher Classifier}) to the Flan-T5 encoder (\emph{Student Classifier}). Specifically, we leverage the outputs $\hat{y}{tc}$ and $\hat{y}{sc}$ from the \emph{Teacher Classifier} and \emph{Student Classifier}, respectively, and employ the Kullback-Leibler divergence loss \cite{csiszar1975divergence} as the \emph{teacher forcing} loss to minimize the discrepancy between $\hat{y}_{tc}$ and $\hat{y}_{sc}$:

\begin{equation}\label{eq:tf-loss} 
\mathcal{L}_{tf} = D_{KL}(\hat{y}_{sc}) || \hat{y}_{tc}))
\end{equation}

\noindent Our final loss $\mathcal{L}$ is the weighted sum of $\mathcal{L}_{xcl}, \mathcal{L}_{tf}$:

\begin{equation}\label{eq:final-loss} 
\mathcal{L} = \lambda\mathcal{L}_{xcl} + \gamma\mathcal{L}_{tf}
\end{equation}

\noindent in which $\lambda, \gamma$ are the contribution weights. 

\subsubsection{Conditional Decoding Constraint (CD)}
\label{sssec:conditional-decoding}
One of the main challenges with unified frameworks for toxic speech detection and explanation is the lack of synchronization between the classifier's output label and the generated explanation. For instance, when the classifier outputs a label of $0$, indicating non-toxic speech, the explanation generation module still generates an explanation, even though it is unnecessary in this case. To address this, we propose the \emph{Conditional Decoding Constraint}, a simple yet effective algorithm. This constraint controls the decoder's generation process by generating a \texttt{[None]} token for non-toxic speech and a toxic explanation for toxic speech. By incorporating this constraint, our framework ensures coherence and alignment between the generated explanations and classifier outputs, enhancing its overall performance.

%%%%%%%%%%%%%%%%%%%%%%%%%%%%%%%%%%%%%%%%%%%%%%%%%%%%%%%%%%%%%%%%%%%%%%%%%%%%%%%%%%%%%%
%%%%%%%%%%%%%%%%%%%%%%%%%%%%%%%%%%%%%%%%%%%%%%%%%%%%%%%%%%%%%%%%%%%%%%%%%%%%%%%%%%%%%%

\section{Experimentation}
\label{sec:experimentation}
\subsection{Target Group Generator Experiment}
\label{ssec:exp-tg-generator}
\noindent \paragraph{$\bullet$ Dataset}
To address the problem of free-text target group labelling in IHC and SBIC datasets, we utilized the \texttt{HateXplain} dataset \cite{hatexplain}, which provides 19 fine-grained categories for toxic speech. We fine-tune a T5 model \cite{2020t5} as our TG model, and treat it as a text generation problem. To ensure our framework can predict specific target group(s) associated with posts from IHC and SBIC datasets, we conducted an analysis to identify any overlapping data between the \texttt{HateXplain} and IHC, SBIC datasets. We found only one instance of overlap, which we removed before training our TG model.

\noindent \paragraph{$\bullet$ Baselines}
We compare the performance of our TG model with three baseline models: (1) BERT \cite{devlin-etal-2019-bert}, an encoder-based model; (2) GPT-2 \cite{radford2019language}, a decoder-only model, (3) and BART \cite{lewis-etal-2020-bart}, an encoder-decoder model. BERT is widely used for multi-label classification tasks while both GPT-2 and BART have demonstrated remarkable performance in text 

\noindent \paragraph{$\bullet$ Implementation Details}
We concatenate the annotated target group(s) in alphabetical order to construct the target label for each input post. All baselines and our TG model are initialized with pre-trained checkpoints from Huggingface \cite{wolf-etal-2020-transformers} and fine-tuned on a single Google CoLab P40 GPU with a window size of 256, a learning rate of $1e-5$, and AdamW \cite{loshchilov2018decoupled} as the optimizer. The BERT model is fine-tuned for 10 epochs, while the GPT-2 and BART models are fine-tuned for 20k iterations. We use a beam search strategy with a beam size of 4 for our generation decoding strategy.

\noindent \paragraph{$\bullet$ Automatic Evaluation}
Our TG model is evaluated using F1 (\%) for multi-label classification and ROUGE-L (\%) \cite{lin-2004-rouge}. The results in Table \ref{tab:target-group-generator} indicate that our model achieved an F1 score of $69.79$ and a ROUGE-L score of $70.95$, outperforming the competing baselines in identifying target groups in toxic posts.

\begin{table}
\centering
\resizebox{0.7\columnwidth}{!}{
    \begin{tabular}{l|cc}
        \toprule 
        Model & F1 & ROUGE-L \\
        \midrule
        BERT & 68.35 & 70.44 \\
        GPT-2 & 53.98 & 56.86  \\
        BART & 63.41 & 70.48  \\
        \midrule
        T5 & \textbf{69.79} & \textbf{70.95}  \\
        \bottomrule
    \end{tabular}
}
\caption{\small{Target Group Generator experiments.}}
\label{tab:target-group-generator}
\end{table}

\subsection{Teacher Classifier Experiment}
\label{ssec:exp-teacher-classifier}
For our Teacher Classifier, a RoBERTa-Large model \cite{liu2019roberta} is fine-tuned using the generated target group(s) (TG in Section~\ref{sec:method}). The model achieved an F1 score of $79.49$ on IHC and $91.19$ on SBIC, indicating the effectiveness of generated target group(s) in classifying toxic speech. Detailed results can be found in Table~\ref{tab:teacher-performance}.

\begin{table}
\centering
\resizebox{0.99\columnwidth}{!}{
    \begin{tabular}{l|c|cc|cc}
    \toprule 
    &  & \multicolumn{2}{c|}{IHC} & \multicolumn{2}{c}{SBIC} \\
    \midrule
    Model & Size & Acc. (\%) & F1 (\%) & Acc. (\%) & F1 (\%) \\
    \midrule
    RoBERTa-Large & 354M & 80.68 & 77.33 & 90.12 & 90.11  \\
    \midrule
    Teacher Classifier & 354M & \textbf{82.52} & \textbf{79.49} & \textbf{91.19} & \textbf{91.19} \\
    \bottomrule
    \end{tabular}
}
\caption{\small{Performance of Teacher Classifier, which is the RoBERTa-Large + \textbf{TG}.}}
\label{tab:teacher-performance}
\end{table}

\subsection{\model{} Experiment}
\label{ssec:exp-toxcl}
\noindent \paragraph{$\bullet$ Dataset}
We conduct our experiments on two datasets: IHC \cite{elsherief-etal-2021-latent} and SBIC \cite{sap-etal-2020-social}. These datasets are collected from popular social media platforms such as Twitter and Gab, providing comprehensive coverage of the most prevalent toxic groups. Prior to training, we pre-process the data as detailed in Appendix~\ref{sec:preprocess-data}.

\noindent \paragraph{$\bullet$ Baselines}
We compare \model{} with three groups of baselines: \emph{(G1) implicit toxic speech detection}, \emph{(G2) implicit toxic speech explanation}, and \emph{(G3) implicit toxic speech detection and explanation}.

For baselines in group G1, we use BERT \cite{devlin-etal-2019-bert}, HateBERT \cite{caselli-etal-2021-hatebert}, RoBERTa \cite{liu2019roberta} and ELECTRA \cite{Clark2020ELECTRA} as our baselines. They are widely employed in prior toxic speech detection works.

We select GPT-2 \cite{radford2019language}, BART \cite{lewis-etal-2020-bart}, T5 \cite{2020t5}, and Flan-T5 \cite{flant5} as our baselines G2 and G3. GPT-2 represents the group of decoder-only models, while BART, T5, and Flan-T5 have the encoder-decoder architecture. Specifically for group G2, we fine-tune them to generate \texttt{[None]} token or the explanations' tokens. For group G3, we concatenate the  classification label \texttt{[Toxic]/[Non-toxic]} and the explanation of each sample as the output, and fine-tune the models with the post as input. Furthermore, to align with recent advancements in Large Language Models (LLMs), we further include ChatGPT \footnote{Version: gpt-3.5-turbo-0613} (a state-of-the-art closed-source LLM) and Mistral-7B-Instruct-v0.2 (a state-of-the-art open-source LLM) in group G3. Both models are evaluated under the zero-shot setting.

\noindent \paragraph{$\bullet$ Implementation Details}
We initialize all the models with the pre-trained checkpoints from Huggingface \cite{wolf-etal-2020-transformers}. We then fine-tune them on a single Google CoLab P40 GPU with a window size of 256, and a learning rate of $1e-5$ and use AdamW \cite{loshchilov2018decoupled} as our optimizer. The classification baselines in group G1 are fine-tuned on $10$ epochs while the generation ones in G2 and G3 are fine-tuned on $20$k iterations. Beam search  strategy with a beam size of $4$ is utilized as our generation decoding strategy.

\noindent \paragraph{$\bullet$ Automatic Evaluation}
We adopt Accuracy and Macro F1 as our classification metrics, following prior works \cite{hatexplain, elsherief-etal-2021-latent}. For the generation of explanations, we utilize BLEU-4 \cite{papineni-etal-2002-bleu}, ROUGE-L \cite{lin-2004-rouge} and METEOR \cite{banerjee-lavie-2005-meteor} as our n-gram metrics. We further utilize BERTScore \cite{bert-score} to measure the similarity between the generated toxic explanation and the ground truth one based on deep-contextual embeddings. To ensure that unnecessary explanations are not generated for non-toxic posts and penalize unnecessary explanations, we develop a new evaluation algorithm for the explanation generation task. Its pseudo-code is presented in Algorithm~\ref{alg:generation-scores}. In this algorithm, we assign a score of $100$ if both the generated explanation and the ground truth explanation are \texttt{[None]} indicating that no explanation is needed. If both the generated explanation and the ground truth explanation are not \texttt{[None]} we compute a score based on the quality of the generated explanation. For any other mismatched pairs, we assign a score of 0 to penalize the unnecessary explanations for non-toxic speech. It is worth noting that our evaluation algorithm is different from \citet{sridhar-yang-2022-explaining} which only evaluates the quality of the generation within implicit toxic cases.

\noindent \paragraph{$\bullet$ Human Evaluation}
To gain deeper insights into the \emph{generation} performance and challenges that our \model{} faces compared to the competing baseline, we conduct human evaluations using a randomly selected set of 150 samples that require explanations from each examined benchmark. Specifically, we collect the generated explanations from both the \model{} and Flan-T5 models in two different settings, G2 and G3. To ensure high-quality evaluations, five native English speakers are hired to rate the generated explanations on a 1-3 scale (with 3 being the highest) based on three criteria: \textbf{(i) Correctness}, evaluating the accuracy of the explanation in correctly explaining the meaning of toxic speech; \textbf{(ii) Fluency}, assessing the fluency and coherence of the generated explanation in terms of language use; and \textbf{(iii) Toxicity}, gauging the level of harmfulness and judgmental tone exhibited in the generated explanation. The annotator agreement is measured using Krippendorff's alpha \cite{krippendorff2011computing}, which provides a measure of inter-annotator reliability.

\subsection{\model{} Performance}
\label{ssec:toxcl-results}

\noindent \paragraph{$\bullet$ Automatic Evaluation}
Our experimental results in Table~\ref{tab:main-experimentations} reveal three main observations. Firstly, \model{} outperforms all baselines on both benchmarks, demonstrating the effectiveness of our encoder-decoder model in addressing both implicit toxic detection and explanation tasks simultaneously without conflicts. Secondly, our model surpasses detection models in group G1, indicating the strong capability of our encoder in detecting implicit toxic speech. It is worth noting that despite having fewer parameters than the RoBERTa-Base model (124M), our encoder (Flan-T5's) classifier (109M) achieves better performance while maintaining computational efficiency. Lastly, our model significantly outperforms its backbone, Flan-T5, highlighting the effectiveness of our architectural designs in jointly training the tasks in an end-to-end manner for implicit toxic speech detection and explanation problems.

\begin{table*}
\centering
\resizebox{0.99\textwidth}{!}{
    \begin{tabular}{l|c|cc|cccc||cc|cccc}
    \toprule
         &  & \multicolumn{2}{c|}{IHC Detection} & 
         \multicolumn{4}{c||}{IHC Explanation} & \multicolumn{2}{c|}{SBIC Detection} & 
         \multicolumn{4}{c}{SBIC Explanation} \\
    \midrule
    Model & Group & Acc. & Macro F1 & BLEU-4 & ROUGE-L & METEOR & BERTScore & Acc. & Macro F1 & BLEU-4 & ROUGE-L & METEOR & BERTScore\\ 
    \midrule
    HateBERT &  & 78.67 & 75.93 & \cellcolor{gray!15} & \cellcolor{gray!15} & \cellcolor{gray!15} & \cellcolor{gray!15} & 89.32 & 89.31 & \cellcolor{gray!15} & \cellcolor{gray!15} & \cellcolor{gray!15} & \cellcolor{gray!15} \\
    BERT & G1 & 78.98 & 76.16 & \cellcolor{gray!15} & \cellcolor{gray!15} & \cellcolor{gray!15} & \cellcolor{gray!15} & 89.83 & 89.83 & \cellcolor{gray!15} & \cellcolor{gray!15} & \cellcolor{gray!15} & \cellcolor{gray!15} \\
    ELECTRA &  & 79.90 & 76.87 & \cellcolor{gray!15} & \cellcolor{gray!15} & \cellcolor{gray!15} & \cellcolor{gray!15} & 89.06 & 89.04 & \cellcolor{gray!15} & \cellcolor{gray!15} & \cellcolor{gray!15} & \cellcolor{gray!15} \\
    RoBERTa  &  & 80.06 & 77.23 & \cellcolor{gray!15} & \cellcolor{gray!15} & \cellcolor{gray!15} & \cellcolor{gray!15} & 89.98 & 89.97 & \cellcolor{gray!15} & \cellcolor{gray!15} & \cellcolor{gray!15} & \cellcolor{gray!15} \\
    \midrule
    GPT-2  &  & \cellcolor{gray!15} & \cellcolor{gray!15} & 67.60 & 70.19 & 69.69 & 73.15 & \cellcolor{gray!15} & \cellcolor{gray!15} & 47.62 & 65.50 & 63.74 & 74.74 \\
    BART  & G2 & \cellcolor{gray!15} & \cellcolor{gray!15} & 53.67 & 59.18 & 57.40 & 68.14 & \cellcolor{gray!15} & \cellcolor{gray!15} & 45.22 & 67.73 & 67.25 & 83.98 \\
    T5 &  & \cellcolor{gray!15} & \cellcolor{gray!15} & 50.19 & 56.01 & 54.36 & 66.60  & \cellcolor{gray!15} & \cellcolor{gray!15} & 45.37 & 68.03 & 67.59 & 84.68 \\
    Flan-T5 &  & \cellcolor{gray!15} & \cellcolor{gray!15} & 47.33 & 53.78 & 51.95 & 65.14 & \cellcolor{gray!15} & \cellcolor{gray!15} & 45.83 & 68.37 & 67.98 & 85.04 \\
    \midrule
    GPT-2  &  & 77.57 & 76.36 & 67.81 & 70.19 & 69.88 & 73.47 & 73.62 & 57.09 & 48.29 & 65.24 & 63.53 & 74.61 \\
    BART  &  & 70.40 & 64.11 & 55.92 & 60.48 & 58.87 & 68.74 & 88.25 & 88.20 & 45.65 & 68.08 & 67.81 & 84.35 \\
    T5 & G3 & 70.71 & 62.95 & 58.24 & 62.42 & 61.18 & 69.30 & 87.55 & 87.48 & 45.92 & 68.37 & 67.82 & 84.78 \\
    Flan-T5 &  & 71.52 & 65.56 & 56.58 & 61.75 & 60.20 & 69.84 & 88.40 & 88.37 & 45.98 & 68.81 & 67.99 & 85.07 \\
    ChatGPT & & 55.83 & 18.62 & 27.30 & 29.15 & 31.16 & 51.40 & 78.59 & 31.38 & 0.36 & 4.32 & 3.36 & 48.7 \\
    Mistral-7B & & 67.36 & 26.31 & 0.29 & 2.31 & 3.39 & 27.11 & 78.36 & 52.03 & 0.58 & 8.21 & 2.19 & 79.88 \\
    \midrule
    GPT-2 + \textbf{TG} & & 74.00 & 58.12 & 67.86 & 70.38 & 69.71 & 73.11 & 78.87 & 77.93 & 48.56 & 65.48 & 63.65 & 75.80 \\
    BART + \textbf{TG} & & 74.28 & 58.88 & 67.75 & 68.17 & 69.68 & 73.67 & 88.55 & 88.52 & 45.70 & 68.41 & 67.81 & 85.03 \\
    T5 + \textbf{TG} & & 76.48 & 64.86 & 67.07 & 70.98 & 70.24 & 73.68 & 87.78 & 87.71 & 46.52 & 68.67 & 67.98 & 85.08 \\
    Flan-T5 + \textbf{TG} & & 78.47 & 70.13 & 65.77 & 70.03 & 69.98 & 75.18 & 88.73 & 88.72 & 47.05 & 68.75 & 68.05 & 85.25 \\
    \midrule
    \midrule
    Flan-T5 + \textbf{CLH} & & 77.16 & 73.67 & 62.15 & 64.36 & 62.21 &  63.94 & 89.19 & 89.19 & 45.88 & 67.24 & 67.63 & 84.96  \\
    Flan-T5 + \textbf{CLH} + \textbf{TG} & & 78.68 & 75.77 & 64.99 & 67.87 & 66.24 &  72.11 & 89.6 & 89.6 & 47.24 & 67.94 & 67.80 & 85.34 \\
    Flan-T5 + \textbf{CLH} + \textbf{TG} + \textbf{TF} & & \textbf{81.53}$\dagger$ & \textbf{78.19}$\dagger$ & 66.49 & 69.11 & 67.52 &  72.14 & \textbf{90.09}$\dagger$ & \textbf{90.08}$\dagger$ & 47.85 & 68.93 & 68.16 & 85.58 \\
    \emph{\model} & & \textbf{81.53}$\dagger$ & \textbf{78.19}$\dagger$ & \textbf{68.11}$\dagger$ & \textbf{71.21}$\dagger$ & \textbf{70.27}$\dagger$ &  \textbf{77.38}$\dagger$ & \textbf{90.09}$\dagger$ & \textbf{90.08}$\dagger$ & \textbf{49.03}$\dagger$ & \textbf{69.93}$\dagger$ & \textbf{68.85}$\dagger$ & \textbf{86.09}$\dagger$ \\
    \bottomrule
    \end{tabular}
}
\caption{\small{Main experimental results. \textbf{CLH} stands for joint training with a classification head on top of the Flan-T5 Encoder. Our model, \model{} is equivalent to Flan-T5 + \textbf{CL Head} + \textbf{TG} + \textbf{TF} + \textbf{CD} (all are described in Section~\ref{ssec:toxcl}). $\dagger$ denotes our model significantly outperforms \emph{implicit toxic speech detection \& explanation} baselines with p-value < 0.05 under t-test.}}
\label{tab:main-experimentations}
\end{table*}

\noindent \paragraph{$\bullet$ Human Evaluation}
Our human evaluation results in Table~\ref{tab:human-evaluation} indicate that \model{} outperforms its backbone model, Flan-T5, from groups G2 and G3, in terms of both explanation accuracy and textual fluency. This improved performance is also reflected in the detection task, resulting in more reliable explanations with fewer harmful outputs compared to the baselines. Our human annotators exhibit strong agreement with Krippendorff’s alpha scores consistently measuring at least 0.78 among the three scores. It is worth noting that the average toxicity score of around 2 for both our \model{} and baseline models aligns with expectations, given that the training datasets contain offensive words in the ground truth explanations (see Tab.~\ref{tab:qualitative-results}). 

\begin{table}
\centering
\resizebox{0.85\columnwidth}{!}{
    \begin{tabular}{l|ccc}
    \toprule 
    Model & Cor.$\uparrow$ & Flu.$\uparrow$ & Tox.$\downarrow$ \\
    \midrule
    Flan-T5 (G2) & 2.21 & 2.02 & 2.03 \\
    Flan-T5 (G3) & 2.35 & 2.46 & 2.07  \\
    \midrule
    \model{}  & \textbf{2.56} & \textbf{2.63} & \textbf{1.97} \\
    \midrule
    \midrule
    Kripp.'s alpha & 0.81 & 0.84  & 0.78  \\
    \bottomrule
    \end{tabular}
}
\caption{\small{Human evaluation results.}}
\label{tab:human-evaluation}
\end{table}

    % Flan-T5 (G2) & 1.81 & 2.02 & 2.03 \\
    % Flan-T5 (G3) & 1.95 & 2.46 & 2.07  \\
    % \midrule
    % \model{}  & \textbf{2.04} & \textbf{2.63} & \textbf{1.97} \\
%%%%%%%%%%%%%%%%%%%%%%%%%%%%%%%%%%%%%%%%%%%%%%%%%%%%%%%%%%%%%%%%%%%%%%%%%%%%%%%%%%%%%%
%%%%%%%%%%%%%%%%%%%%%%%%%%%%%%%%%%%%%%%%%%%%%%%%%%%%%%%%%%%%%%%%%%%%%%%%%%%%%%%%%%%%%%

\section{Discussion}
\label{sec:discussion}

\subsection{Case Study: Effectiveness of \model{}}
\label{ssec:case-study}
We compare the performance of \model{} with the two best-performing baselines in group G3, namely GPT-2 on IHC and Flan-T5 on SBIC (Sec.~\ref{sec:experimentation}). We present the cases discussed in Table~\ref{tab:qualitative-results}. In cases (1), (3), (4), (7), (9), and (10) we observe that both GPT-2 and Flan-T5 fail to capture the intended meaning of the input posts, resulting in wrong detection or harmful explanations. In contrast, our \model{} effectively captures the meaning of the posts, leading to accurate explanations that align closely with the ground truth ones. Notably, when comparing to the ground truth explanation, the explanation generated by \model{} exhibits a more polite attitude (cases 2, 4, 8, 9) or even more accurate (case 3).

\begin{table*}
\centering
\resizebox{0.99\textwidth}{!}{
    \begin{tabular}{p{0.7cm}|p{7.2cm}p{4.75cm}|p{2.35cm}p{2.45cm}|p{2.5cm}}
    \toprule
    Index & Input Post & Ground Truth Explanation(s) & GPT-2 (IHC) & Flan-T5 (SBIC) & \model{} \\ 
    \hline
    (1) & approx half muslms today result of inbreeding, perhaps this why they are ""hell bent"" on destroying christian based west? inbreed = low iq. & Muslims are result of inbreeding & Muslims are destructive & \cellcolor{gray!15} & Muslims are inbred.  \\
    \midrule
    (2) & you hate like a jew. you are fixatd on hating white christians like a jew. & Jewish people have hate toward other groups & Jews hate white people & \cellcolor{gray!15} & Jews don't like white poeple \\
    \midrule
    (3) & funny how you never hear about white men raping women in black" neighborhoods. i know 2 white women who ’ & Blacks rape whites. & Blacks are inferior & \cellcolor{gray!15} & White men are rapists. \\
    \midrule
    (4) & she's another low iq hollywood liberal . islam is submission and btw if they ran america & Liberals are stupid. & None & \cellcolor{gray!15} & Liberals are low IQ people. \\
    \midrule
    (5) & rt. : " voters on obamas \#immigration plans : drop dead " " \#noamnesty " & Immigrants ruin society & None & \cellcolor{gray!15} & None \\
    \midrule
    (6) & bruh i love white people too how do i join & Minorities are lesser races. & None & \cellcolor{gray!15} & None \\
    \midrule
    \midrule
    (7) & An Asian woman just gave me a handjob while fingering my ass Which isnt what i meant when i asked her to give me a 'Hand Roll and Pork Intestines. That's the problem with having the Chinese Massage place right next to the Chinese takeout place & ['chinese women are sex workers.', "chinese women don't understand english."] & \cellcolor{gray!15} & asians eat pork & asian women only care about sex \\
    \midrule
    (8) & you see, i am a doctor. today i had to save a black guy's life i told his family: i did everything i could, but unfortunately he survived. & ['black people are useless', 'black people need to be killed'] & \cellcolor{gray!15} & black people are worthless & black people aren't as good as whites \\
    \midrule
    (9) & @BlakkHoodie\_562 yeah that niggah played good, lowkey i wanted brazil to win & ['black people are the n word'] & \cellcolor{gray!15} & None & niggah is a slang term \\
    \midrule
    (10) & A bf and gf; are suppose to laugh; at these hoes; together;. A guy should never have these hoes; laughing; at his girl; & ['degrades women by calling them names', 'women are called hoes', 'women are hoes'] & \cellcolor{gray!15} & None & women are hoes \\
    \midrule
    \bottomrule
    \end{tabular}
}
\caption{\small{The qualitative explanation performance of GPT-2, Flan-T5, and \model{} was evaluated on samples from the valid set of IHC and SBIC datasets.}}
\label{tab:qualitative-results}
\end{table*}

\subsection{Ablation Studies}
\label{ssec:ablation-studies}

\paragraph{$\bullet$ Target Group Generator (TG)}
When adding the this module (i.e., + \textbf{TG} in Table~\ref{tab:main-experimentations}), we observe that all baselines considered in our experiments achieve significant improvements compared to training them without the generated target groups. In addition, our Teacher Classifier also outperforms its variant trained on solely the input posts. These improvements emphasize the crucial role of the target group(s) in enhancing their ability to detect and explain implicit toxic speech. These results also consolidate our motivation for using generated target groups in guiding the models for solving the two tasks.

\paragraph{$\bullet$ Teacher Forcing (TF)}
By comparing the performance of Flan-T5 + \textbf{CLH} + \textbf{TG} before and after using \emph{teacher forcing} (+ \textbf{TF} in Table~\ref{tab:main-experimentations}), we observe that incorporating this technique improves the performance of the encoder of \model{}. This improvement demonstrates that providing additional guidance to the encoder of \model{} results in more accurate predictions and achieves a performance close to that of the \emph{Teacher Classifier}.

\paragraph{$\bullet$ Conditional Decoding Constraint (CD)}
Finally, the impact of integrating the \emph{Conditional Decoding Constraint} designed to avoid necessary explanation, is evaluated in Table~\ref{tab:main-experimentations}. Compared with Flan-T5 + \textbf{CLH} + \textbf{TG} + \textbf{TF}, our \model{} is improved significantly on the toxic explanation generation task. This confirms the effectiveness of \emph{Conditional Decoding Constraint} in helping the outputs of our model to synchronize implicit toxic speech detection labels and toxic explanations.

\subsection{Error Analysis}
\label{ssec:error-analysis}
We present additional qualitative samples from both benchmarks in Table~\ref{tab:qualitative-results} to highlight key challenges faced by existing models in detecting implicit toxic speech and generating explanations. While our model performs well overall, there are still areas for improvement, as discussed below.

\paragraph{$\bullet$ Coded Toxic Symbols}
Our model, along with the baselines, struggles with detecting implicit toxic speech that contains abbreviated or coded tokens, such as "\emph{\#noamnesty"} in case (5). This error has also been observed and discussed by previous work \cite{elsherief-etal-2021-latent}.

\paragraph{$\bullet$ Misunderstood Sarcasm and Irony}
Our model may face challenges in accurately detecting toxic sentences that contain indirect words. For example, case (6) involves the phrase \emph{"bruh i love white people too how do i join"}, which uses indirect words such as "bruh" and "love" to express irony. The speaker sarcastically expresses a desire to join a racial group while implying that joining such a group is based on a belief in the superiority of white people or that minorities are lesser races.

\paragraph{$\bullet$ Variant Explanations}
In all cases except cases (5), (6), and (10), our \model{} accurately identifies implicit toxic speech but generates linguistically different explanations from the ground truth(s). However, these generated explanations convey the same semantic meaning as the ground truth, indicating the model's ability to comprehend correctly the implicit meanings. This, along with discussions by \citet{huang2022chain}, demonstrates that instances of implicit toxic speech can have multiple correct explanations, highlighting the limitations of commonly-used n-gram evaluation metrics like BLEU-4 \cite{papineni-etal-2002-bleu} and ROUGE-L \cite{lin-2004-rouge} scores.

%%%%%%%%%%%%%%%%%%%%%%%%%%%%%%%%%%%%%%%%%%%%%%%%%%%%%%%%%%%%%%%%%%%%%%%%%%%%%%%%%%%%%%
%%%%%%%%%%%%%%%%%%%%%%%%%%%%%%%%%%%%%%%%%%%%%%%%%%%%%%%%%%%%%%%%%%%%%%%%%%%%%%%%%%%%%%

\section{Conclusion}
\label{sec:conclusion}
We present \model{}, a unified framework for implicit toxic speech detection and explanation (Sec.~\ref{ssec:toxcl}). It consists of three components: a Target Group Generator, an Encoder-Decoder model, and a Teacher Classifier. Our findings show that the Target Group Generator effectively identifies target groups, improving both accuracy and F1 scores for detecting implicit toxic speech. The novel encoder-decoder architecture successfully performs both tasks of detection and explanation without harming each other. The integration of the Teacher Classifier and the Conditional Decoding Constraint further enhances the performance of \model{},  achieving state-of-the-art results in the task of toxic speech detection and explanation on two widely-used benchmarks. In the future, we will focus on addressing several limitations faced by our framework and baselines as specified in Section~\ref{ssec:error-analysis} to further enhance the performance of our model.

%%%%%%%%%%%%%%%%%%%%%%%%%%%%%%%%%%%%%%%%%%%%%%%%%%%%%%%%%%%%%%%%%%%%%%%%%%%%%%%%%%%%%%
%%%%%%%%%%%%%%%%%%%%%%%%%%%%%%%%%%%%%%%%%%%%%%%%%%%%%%%%%%%%%%%%%%%%%%%%%%%%%%%%%%%%%%

\section*{Limitations}
\label{sec:limitations}
Although our model has demonstrated strong performance, our error analysis (Sec.~\ref{ssec:error-analysis}) has identified several challenges that still need to be addressed. One such challenge is the presence of coded toxic symbols, abbreviated words, or implicit phrases in the input posts, which may require external sources of knowledge for accurate interpretation. To address this, future work can focus on enhancing the models by incorporating additional reasoning capabilities and leveraging external knowledge. Additionally, existing evaluation metrics for the implicit toxic speech explanation task is also another limitation since this can be a one-to-many relationship problem in which there may have multiple non-overlapping but correct explanations for an implicit hate speech.

%%%%%%%%%%%%%%%%%%%%%%%%%%%%%%%%%%%%%%%%%%%%%%%%%%%%%%%%%%%%%%%%%%%%%%%%%%%%%%%%%%%%%%
%%%%%%%%%%%%%%%%%%%%%%%%%%%%%%%%%%%%%%%%%%%%%%%%%%%%%%%%%%%%%%%%%%%%%%%%%%%%%%%%%%%%%%

\section*{Ethical Considerations}
While our method for implicit toxic speech has shown promise in identifying target groups, detecting implicit toxic speech, and generating explanations, it is crucial to consider the potential risks involved.

Firstly, there is a concern that the generated explanations may contain toxic words, depending on the training data (Sec.~\ref{tab:qualitative-results}). This raises the possibility of the model spreading machine-generated toxic speech if it is misused. It is essential to address this toxicity to protect marginalized groups and shift power dynamics to the targets of oppression.

Secondly, there is a risk of reinforcing biases or amplifying harmful messages by providing explanations only for detected implicit toxic speech. If the model fails to detect implicit toxic speech, the absence of an explanation may imply acceptability or harmlessness. Considering explanations for all posts, regardless of detection, could be an approach to mitigate this risk, although our datasets do not provide explanations for non-toxic speech.

In conclusion, while having complete control over the \model{}'s usage in real-world scenarios may not be feasible, it is essential to recognize and address potential risks. By doing so, our work offers an opportunity to combat harm directed at minority groups and empower targets of oppression. Therefore, addressing the quality of explanations in the training data is a critical step toward achieving our ultimate goal.

\section*{Acknowledgements} This research/project is supported by the National Research Foundation, Singapore under its AI Singapore Programme, AISG Award No: AISG2-TC-2022-005. Do Xuan Long is supported by A*STAR Computing and Information Science (ACIS) Scholarship.

%%%%%%%%%%%%%%%%%%%%%%%%%%%%%%%%%%%%%%%%%%%%%%%%%%%%%%%%%%%%%%%%%%%%%%%%%%%%%%%%%%%%%%
%%%%%%%%%%%%%%%%%%%%%%%%%%%%%%%%%%%%%%%%%%%%%%%%%%%%%%%%%%%%%%%%%%%%%%%%%%%%%%%%%%%%%%

\bibliography{acl}

\appendix

\section{Data Pre-process}
\label{sec:preprocess-data}
To facilitate our problem, we exclude instances of implicit toxic speech that lack an explanation, resulting in the removal of 844 samples from IHC and 8,220 samples from SBIC. Since the original IHC dataset does not include a designated test set, we create our own by randomly selecting 20\% of the implicit toxic speech and non-toxic speech instances. The final statistics of both datasets are shown in Table~\ref{tab:dataset-statistics}.

\begin{table}[ht!]
\centering
\resizebox{0.8\columnwidth}{!}{
    \begin{tabular}{l|c|c|c}
        \toprule
        Split  & \# toxic & \# non-toxic  & \# samples\\
        \midrule
        IHC Train & 5,002 & 10,633 &  15,635 \\
        IHC Valid & 1,254 & 2,658  & 3,912\\
        \midrule
        SBIC Train & 12,098 & 16,698 & 28,796 \\
        SBIC Dev & 1,806 & 2,054 & 3,860 \\
        SBIC Test & 1,924 & 1,981 & 3,905 \\
        \bottomrule
    \end{tabular}
}
\caption{Statistics of \texttt{Implicit Hate Corpus} and \texttt{Social Bias Inference Corpus} after pre-processing.}
\label{tab:dataset-statistics}
\end{table}

\section{Evaluation Algorithm}

We present our evaluation metrics to evaluate the explanation generation capability of the models. To penalize unnecessary explanations for non-toxic speech, we add 100 to every score when the label is "None" and the model generation output is "None". 

\begin{algorithm}[ht!]
\DontPrintSemicolon
\caption{Computations of Explanation Generation Metrics} \label{alg:generation-scores}
\textbf{Input: } $b\_gen\_labels$, $b\_gen\_expls$ \\
\textbf{Initialize: } \\
$N$ = len($b\_gen\_labels$) \\
$s\_bleu$ = $s\_rouge$ = $s\_meteor$ = $s\_bertscore$ = 0 \\

\For{$idx$ \textbf{in} range($N$)}{
    label = $b\_gen\_labels[idx]$ \\
    expl = $b\_gen\_expls[idx]$ \\
    \If{$label$ == \texttt{"None"} \textbf{and} $expl$ == \texttt{"None"}}{
        add $100$ to $s\_bleu$, $s\_rouge$, $s\_meteor$, $s\_bertscore$ \\
    }
    \ElseIf{$label$ $\neq$ \texttt{"None"} \textbf{and} $expl$ $\neq$ \texttt{"None"}}{
        $s\_bleu$ += bleu($label$, $expl$) \\
        $s\_rouge$ += rouge($label$, $expl$) \\
        $s\_meteor$ += meteor($label$, $expl$) \\
        $s\_bertscore$ += bertscore($label$, $expl$) \\
    }
}

\Return $s\_bleu / N$, $s\_rouge / N$, $s\_meteor / N$, $s\_bertscore / N$
\end{algorithm}

\end{document}